\documentclass{article}

\usepackage[T1]{fontenc}
\usepackage[utf8]{inputenc}
\usepackage{microtype}
\usepackage{graphicx}
\usepackage{booktabs}
\usepackage{hyperref}
\usepackage[preprint]{icml2026}
\usepackage{amsmath,amssymb,mathtools}
\usepackage{tabularx}
\usepackage{array}
\usepackage{url}
\usepackage{tikz}
\usetikzlibrary{arrows.meta}

\newcolumntype{L}[1]{>{\raggedright\arraybackslash}p{#1}}
\newcommand{\tg}{\ensuremath{T_b(\epsilon)}}
\newcommand{\frontier}{\ensuremath{\varphi_b(t)}}

\newcommand{\artifact}[1]{\url{#1}}
\DeclareMathOperator*{\argmin}{arg\,min}

\icmlkeywords{Bayesian inference, decision theory, AI evaluation archives, selective reporting, frontier validation}

\begin{document}

\twocolumn[
\icmltitle{Bayesian Inference and Decision Audits for Public Archives of Frontier AI Evaluations}

\begin{icmlauthorlist}
\icmlauthor{Yanan Long}{yl}
\end{icmlauthorlist}
\icmlaffiliation{yl}{StickFlux Labs}
\icmlcorrespondingauthor{Yanan Long}{ylong@uchicago.edu}

]

\printAffiliationsAndNotice{}

\begin{abstract}
Public AI evaluations are often read as terminal leaderboards, yet the underlying evidence is a selective time series shaped by reporting rules, benchmark revisions, and missingness. Repeated public archives for LiveBench and Open LLM Leaderboard v2 serve as the primary longitudinal record; LMArena provides a preference stress test; and GAIA and tau-bench contribute limited agentic pilots. Together, these archives instantiate a Bayesian inference problem: under a fixed reporting convention, one constructed terminal-only example over $1{,}000$ systems is compatible with two pre-terminal histories, yielding times of $23.03$ or $75.13$ to reach within $0.05$ of the ceiling under the same terminal-tail model. In synthetic posterior comparisons, action-facing diagnostics differ across observation regimes. The candidate selection-aware frontier model fails synthetic recovery, objective-archive prediction, preference transfer, and uncertainty calibration; correspondingly, fixed audit gates reject its stronger claims. An archive-and-adjudication protocol reconstructs public evaluation histories, isolates a verified timing boundary, and falsifies unsupported frontier claims.
\end{abstract}


\section{Introduction}

Public evaluation reporting for modern AI systems is selective and time-indexed: a leaderboard, benchmark release, arena slice, or agentic task report usually exposes a ranked or top-$k$ subset while hiding attempted, unpublished, revised, scaffolded, or under-threshold systems. Agentic reports may also aggregate over prompts, tools, planners, judges, and environment settings without exposing the execution trace. A terminal ranking is therefore a lossy endpoint of a repeated public history, not the history itself.

That history is the primary technical object here. Repeated snapshots retain report timing, benchmark version, visibility rule, and the metadata needed to distinguish evidential change from reporting artifact. We focus on public AI evaluation histories reconstructable from source-native releases, including objective benchmark records, preference leaderboards, and aggregate agentic-result reports. An archive contract, an observability boundary, and an adjudication protocol together make repeated public evaluation records analyzable under selection and benchmark drift: the contract specifies what metadata a snapshot must carry, the boundary separates what is recoverable from repeated traces from what requires full execution logs, and the protocol adjudicates candidate inference methods against held-out observations.

The boundary question sits at the intersection of selective inference and winner's-curse correction \citep{zrnic2025winner}, operational frontier estimation \citep{liu2022quasi,einmahl2025ultimate}, and plateau identification in cure-style settings \citep{jackson2025plateau,yuen2026sufficient}. The specific evidential property that distinguishes this setting is that repeated selected records are the primary object, archive construction is part of the scientific claim, and stronger model statements must pass fixed audit gates rather than borrowing credibility from a terminal rank.

We therefore separate archive-level evidence from model endorsement. Repeated public histories can be reconstructed, graded, compared across observation regimes, and used to adjudicate candidate inference methods. The selection-aware frontier model applied in this work is evaluated against recovery, predictive, transfer, and calibration criteria; the protocol is designed to distinguish cases where those criteria are met from cases where they are not, rather than absorbing unsupported frontier-inference claims into aggregate leaderboard narratives. The current candidate meets none of those four criteria, establishing the scope of what this archive and protocol can certify at present.

The evidence spans two primary objective archives, one preference stress test, and two agentic applicability pilots. GAIA \citep{mialon2023gaia} provides a bounded aggregate public-result history for general-assistant agentic tasks, and tau-bench \citep{yao2024taubench} provides the same for tool-agent-user interaction; both pilots contribute applicability evidence but neither supplies full agent-trace observability. Repeated snapshots are normalized into a graded archive; the observability boundary between repeated traces and terminal records is then verified formally; Bayesian posterior expected-loss diagnostics test action sensitivity within that boundary; and fixed audit gates adjudicate the candidate selection-aware frontier model against each criterion in turn.

\section{Related Work}

AI evaluation work has repeatedly exposed the limits of terminal scores and static leaderboards. Prior work argues that final test numbers should be accompanied by search and validation traces \citep{dodge2019show}, that leaderboard rank can diverge from user utility \citep{ethayarajh2020utility}, and that small benchmark differences are often statistically fragile \citep{card2020power}. Benchmark-repair efforts push in the same direction by emphasizing validity, headroom, dynamic data collection, diagnostic leaderboards, holistic multi-metric reporting, contamination checks, sensitivity audits, and explicit saturation analysis \citep{bowman2021fix,kiela2021dynabench,liu2021explainaboard,liang2023helm,sainz2023contamination,alzahrani2024targets,akhtar2026plateau}. Continuously updated benchmarks such as LiveBench \citep{white2024livebench} and Open LLM Leaderboard v2 \citep{openllmleaderboardv2} address some contamination and maintenance issues, but supporting temporal archive claims requires source-native histories that these formats do not currently provide.

The statistical concern also overlaps with adaptive data analysis, reliable leaderboard mechanisms, post-selection inference, and winner's-curse correction \citep{dwork2015adaptive,blum2015ladder,roelofs2019meta,berk2013postselection,andrews2024winners,zrnic2025winner}. Those literatures explain why selected frontier records should not be interpreted as ordinary fixed estimates. Addressing this gap requires reconstructing source-native public traces, marking missing observations, and testing candidate inference methods against future observations rather than treating the observed winner as an unbiased target. An archive contract with fixed audit gates applied to selected public records provides the structure needed to operationalize these statistical requirements in practice.

Several baseline families are natural comparators for this archive setting. Item-response and latent-ability models for evaluation leaderboards treat examples as unequally informative and separate item difficulty from system skill \citep{lalor2016irt,rodriguez2021examples,vania2021irt}. Scaling-law and capability-extrapolation work provides a different frontier baseline while also warning that aggregate metrics can induce or hide apparent discontinuities \citep{kaplan2020scaling,hoffmann2022compute,srivastava2023bigbench,schaeffer2023mirage}. Preference leaderboards inherit Bradley-Terry and Elo-style assumptions \citep{bradley1952rank,elo1978rating}, and modern Arena-style evaluations add prompt, population, judge, and release-timing confounds \citep{zheng2023mtbench,chiang2024arena}. Together, these baselines motivate treating latent-effect/factor, terminal-history, rolling-frontier, and Bradley-Terry / Elo methods as comparators instead of assuming a selection-aware model should win.

Agentic benchmarks extend evaluation from base models to system configurations involving tools, browsing, code execution, environments, simulators, and judges. GAIA and tau-bench are central examples for general-assistant and tool-agent-user interaction evaluation \citep{mialon2023gaia,yao2024taubench}, while SWE-bench, WebArena, AgentBench, and ToolLLM illustrate the broader move toward execution-grounded and tool-mediated tasks \citep{jimenez2024swebench,zhou2024webarena,liu2024agentbench,qin2024toollm}. These benchmark families reveal the metadata requirements that any temporal archive covering agentic tasks must satisfy; the GAIA and tau-bench pilots represent early archive scope rather than primary evidence for frontier model ranking.

\section{Evaluation-Trace Archive}

Each source record stores the public source, snapshot unit, timestamp field, score fields, score orientation, rank handling, duplicate policy, missingness summary, and inclusion grade. Scores are converted to a single higher-is-better orientation, and duplicates collapse to one canonical record per source, benchmark, timestamp, system, and task group, preferring explicit rank, then better rank, then higher score. Timestamps must be source-native or explicitly flagged as derived, and any source that fails schema, timestamp, orientation, duplicate, top-$k$ reconstruction, or role-specific minimum-history checks is excluded from the corresponding evidence role rather than silently entering downstream tables.

Table~\ref{tab:source-validation} gives the compact source-validation readout used in the main text. LiveBench \citep{white2024livebench} and Open LLM Leaderboard v2 \citep{openllmleaderboardv2} are the primary objective archives, while LMArena \citep{chiang2024arena} serves as a preference stress test. GAIA is included as a secondary agentic pilot with $463$ snapshots, $3{,}353$ systems, and $11{,}784$ diagnostic rows, and tau-bench is included as an agentic stress-test pilot with $10$ snapshots, $27$ systems, and $27$ diagnostic rows; diagnostic rows are pilot result rows, whereas validation slices are rolling-origin fold counts. LiveCodeBench, HELM Capabilities \citep{liang2023helm}, and SWE-bench Verified \citep{jimenez2024swebench} remain excluded because the public histories used here did not provide the versioned source tables needed for temporal reconstruction. For agentic benchmarks, the archive contract must also record tool budget, environment version, scaffold identity, retry policy, judge version, and human-intervention policy to support temporal reconstruction.

\begin{table*}[t]
\centering
\small
\setlength{\tabcolsep}{3.5pt}
\caption{Compact source-validation readout for the public evaluation archive. Roles define how each source can be used in the manuscript evidence: primary objective prediction, preference stress testing, agentic applicability, or explicit exclusion.}
\label{tab:source-validation}
\begin{tabular}{L{0.23\textwidth}L{0.19\textwidth}L{0.22\textwidth}L{0.27\textwidth}}
\toprule
Source & Public role & Validated scale & Use \\
\midrule
LiveBench & Primary objective archive & $94$ snapshots; $195$ systems & Primary rolling-origin objective prediction evidence. \\
Open LLM Leaderboard v2 & Primary objective archive & $262$ snapshots; $4{,}484$ systems & Primary rolling-origin objective prediction evidence. \\
LMArena & Preference stress test & $152$ snapshots; $365$ systems & Separate Arena-style stress test; not an objective archive. \\
GAIA & Secondary agentic pilot & $463$ snapshots; $3{,}353$ systems; $11{,}784$ diagnostic rows & Archive-applicability evidence for aggregate agentic histories. \\
tau-bench & Agentic stress-test pilot & $10$ snapshots; $27$ systems; $27$ diagnostic rows & Archive-applicability evidence for tool-use submission histories. \\
LiveCodeBench; HELM Capabilities; SWE-bench Verified & Excluded & Not counted in the evidence baseline & Public histories used here lacked the versioned source tables needed for temporal reconstruction. \\
\bottomrule
\end{tabular}
\end{table*}

\paragraph{Archive validation protocol.}
\texttt{main} denotes eligibility for the primary objective rolling-origin backtest, \texttt{stress-test} denotes use only in a source-specific stress regime, \texttt{secondary} denotes an archive-applicability pilot outside the main objective evidence, and \texttt{excluded} denotes a source retained in the manifest but not counted in the evidence baseline. To qualify as \texttt{main}, a source must be an objective score archive and must have at least $5$ validated snapshots, at least $10$ distinct canonical systems, at least $3$ eligible one-step folds, and at least $1$ eligible two-step fold; \texttt{stress-test} and \texttt{secondary} sources must have at least $3$ validated snapshots but cannot support the primary objective headline unless formally regraded as \texttt{main}. Eligible fold counts are deterministic functions of the validated snapshot history. Index validated snapshots in time order as $1,\ldots,n_{\mathrm{snap}}$. A one-step rolling-origin fold trains on snapshots $1{:}j$ and forecasts snapshot $j+1$; the admissible one-step origins are $j=3,\ldots,n_{\mathrm{snap}}-1$, giving $n_{\mathrm{snap}}-3$ folds when positive. A two-step rolling-origin fold trains on snapshots $1{:}j$ and forecasts snapshots $j+1{:}j+2$; the admissible two-step origins are $j=3,\ldots,n_{\mathrm{snap}}-2$, giving $n_{\mathrm{snap}}-4$ folds when positive. Diagnostic-admissible rows are the fold rows that survive archive validation, comparator availability, and fold-eligibility checks, and only those rows may enter the objective backtest and calibration audit. Together with the row-level checks above and the manifest release pointers in Appendix~\ref{app:tables}, these thresholds make every inclusion decision reproducible from the public archive.

\section{Evaluation-Trace Observability Regime}

For evaluation source $b$ at reporting time $t$, we treat the public snapshot as a selected record: it contains a selected set $S_{bt}$ with observed scores or preference summaries $\{y_{bti}: i \in S_{bt}\}$, a reported cutoff size $k_{bt}$ when the source exposes one, and auxiliary archive metadata $a_{bt}$. For simple model leaderboards, $i$ indexes a model submission; for richer agentic evaluations, it may index a system configuration that includes the base model, prompt, tools, memory, planner, environment policy, and judge. The repeated-snapshot archive is therefore
\begin{align}
D_b^{\mathrm{snap}} = \{(t, S_{bt}, y_{bti}: i \in S_{bt}, k_{bt}, a_{bt})\}_{t \in \mathcal{T}_b}.
\end{align}
The metadata $a_{bt}$ is part of the evidence rather than bookkeeping because it records benchmark version, source timestamp, score orientation, rank handling, duplicate policy, task slice, and any available evaluator or environment information.

A terminal-only archive keeps only the final selected public record,
\begin{align}
D_b^{\mathrm{term}} = (t_T, S_{bT}, y_{bTi}: i \in S_{bT}, k_{bT}, a_{bT}).
\end{align}
The contrast is consequential for identification. The repeated archive retains a time-indexed sequence of selected measurements together with their public reporting context, whereas the terminal archive compresses that sequence to a single selected cross-section. Terminal records can therefore support terminal-rank or terminal-score summaries, but they discard the temporal evidence needed for claims about how headroom changed before the final report.

For frontier-style questions, the target is a path functional defined under a fixed reporting convention rather than the realized winner in the observed candidate mix. After fixing the score orientation, reference scale, and population or pool size against which the frontier is evaluated, let $F_{bt}^{\star}$ denote the source-normalized score law at time $t$ under that convention. Here $F_{bt}^{\star}$ and $u_b$ are convention-defined targets on the normalized scale, not directly observed quantities. One useful standardized frontier is
\begin{align}
\frontier = (F_{bt}^{\star})^{-1}(1 - 1 / m^\star),
\end{align}
with fixed reference pool size $m^\star$. The boundary construction below uses $m^\star=m=1000$. The corresponding headroom path is
\begin{align}
\delta_b(t) = u_b - \frontier,
\end{align}
and the timing functional is
\begin{align}
\tg = \inf \{ t : \delta_b(t) \le \epsilon \},
\end{align}
where $u_b$ is the source-normalized score ceiling convention and $\epsilon$ is a residual-gap target on that same normalized scale. The boundary construction fixes $u_b=1$ and $\epsilon=0.05$ as conventions; the resulting $T_b$ values are conditional on that choice rather than universal constants. Timing is not an independently observed event: it inherits the information and limitations of the trace used to infer the headroom path. Other evaluation-trace targets include future-observation prediction, benchmark saturation, rank stability, preference drift, judge sensitivity, and action stability under candidate deployment decisions.

\emph{Identification boundary.} Repeated selected records can support timing inference only under a fixed reporting convention: source metadata, reporting rules, system identities, score orientation, and an explicit path convention must link observations across reporting times. Terminal-only records constrain the selected terminal law but do not, by themselves, identify the pre-terminal headroom path under that convention.

The boundary construction uses one archive at $t_T=10$ with shared $(m,k)=(1000,10)$ and a shared generalized-Pareto terminal law on the normalized score scale with $(\mu,\sigma,\xi)=(0.8160602794,0.18,-0.12)$. Path A uses $(\delta_\infty,\delta_0,\nu,\lambda)=(0,0.5,1,0.1)$ and path B uses $(0,0.2246644821,1,0.02)$. Both induce the same terminal gap $0.1839397206$ and therefore the same terminal selected likelihood; the verification artifact records maximum absolute log-density difference $0.0$. The implied timing targets differ materially: \tg\ is $23.03$ for path A and $75.13$ for path B. This construction is one admissible example, not a general identified-set result. It establishes that terminal-only archives can leave pre-terminal timing underdetermined, with the magnitude depending on the reporting rule, pool size, and frontier-family assumptions.

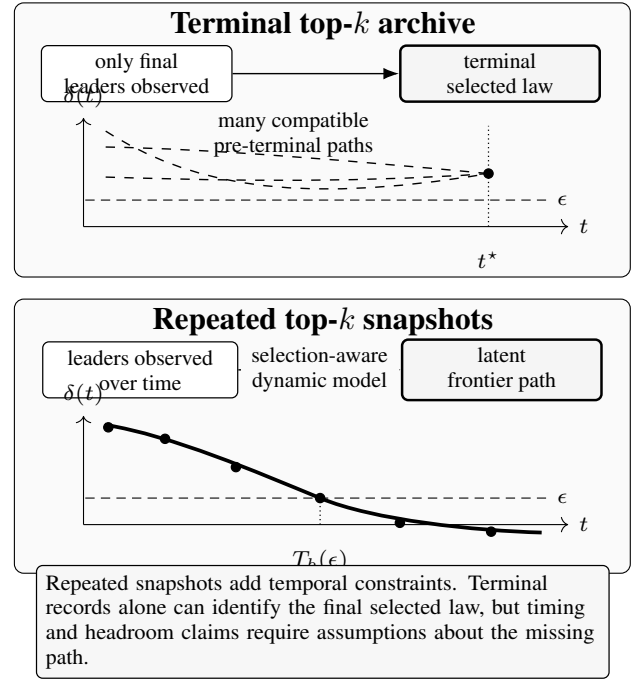
\begin{figure}[t]
\centering
\resizebox{\linewidth}{!}{\begin{tikzpicture}[
  x=1cm,
  y=1cm,
  font=\scriptsize,
  panel/.style={draw, rounded corners=3pt, fill=black!2, line width=0.35pt},
  box/.style={draw, rounded corners=2pt, fill=white, align=center, inner sep=2.6pt},
  target/.style={draw, thick, rounded corners=2pt, fill=black!4, align=center, inner sep=2.6pt},
  note/.style={draw, rounded corners=2pt, fill=black!3, align=left, text width=6.25cm, inner sep=3pt},
  arrow/.style={-Latex, line width=0.45pt},
  faint/.style={line width=0.45pt, dashed},
  axis/.style={->, line width=0.35pt},
  dot/.style={circle, draw=black, fill=black, inner sep=1.05pt}
]

\draw[panel] (0,4.25) rectangle (6.95,7.35);
\node[font=\bfseries] at (3.48,7.10) {Terminal top-$k$ archive};
\node[box, minimum width=2.15cm] (termobs) at (1.38,6.55) {only final\\leaders observed};
\node[target, minimum width=2.25cm] (termlaw) at (5.48,6.55) {terminal\\selected law};
\draw[arrow] (termobs) -- (termlaw);

\draw[axis] (0.78,4.82) -- (6.25,4.82) node[right] {$t$};
\draw[axis] (0.78,4.82) -- (0.78,6.05) node[above] {$\delta(t)$};
\draw[densely dashed, line width=0.35pt] (0.80,5.12) -- (6.00,5.12) node[right] {$\epsilon$};
\draw[densely dotted, line width=0.40pt] (5.35,4.82) -- (5.35,5.96);
\node[below] at (5.35,4.66) {$t^\star$};
\draw[faint] (1.03,5.90) .. controls (1.95,5.32) and (3.10,5.04) .. (5.35,5.42);
\draw[faint] (1.03,5.72) .. controls (2.25,5.70) and (3.85,5.56) .. (5.35,5.42);
\draw[faint] (1.03,5.38) .. controls (2.05,5.35) and (3.70,5.30) .. (5.35,5.42);
\node[dot] at (5.35,5.42) {};
\node[align=center] at (3.15,5.86) {many compatible\\pre-terminal paths};

\draw[panel] (0,0.90) rectangle (6.95,4.00);
\node[font=\bfseries] at (3.48,3.75) {Repeated top-$k$ snapshots};
\node[box, minimum width=2.20cm] (snapobs) at (1.42,3.20) {leaders observed\\over time};
\node[target, minimum width=2.25cm] (path) at (5.50,3.20) {latent\\frontier path};
\draw[arrow] (snapobs) -- node[midway, fill=black!2, align=center] {selection-aware\\dynamic model} (path);

\draw[axis] (0.78,1.45) -- (6.25,1.45) node[right] {$t$};
\draw[axis] (0.78,1.45) -- (0.78,2.68) node[above] {$\delta(t)$};
\draw[densely dashed, line width=0.35pt] (0.80,1.75) -- (6.00,1.75) node[right] {$\epsilon$};
\draw[very thick] (1.02,2.58) .. controls (1.80,2.45) and (2.72,2.05) .. (3.45,1.75)
  .. controls (4.05,1.50) and (5.20,1.38) .. (5.95,1.36);
\foreach \x/\y in {1.06/2.55,1.70/2.42,2.50/2.10,3.45/1.75,4.35/1.47,5.38/1.37} {
  \node[dot] at (\x,\y) {};
}
\draw[densely dotted, line width=0.40pt] (3.45,1.45) -- (3.45,1.75);
\node[below] at (3.45,1.29) {$T_b(\epsilon)$};

\node[note] at (3.48,0.35) {Repeated snapshots add temporal constraints. Terminal records alone can identify the final selected law, but timing and headroom claims require assumptions about the missing path.};

\end{tikzpicture}}
\caption{Repeated top-$k$ snapshots constrain a common latent path. Terminal-only archives constrain only the selected terminal law and can leave pre-terminal timing unidentified.}
\label{fig:observability}
\end{figure}

\begin{figure}[t]
\centering
\includegraphics[width=0.9\linewidth]{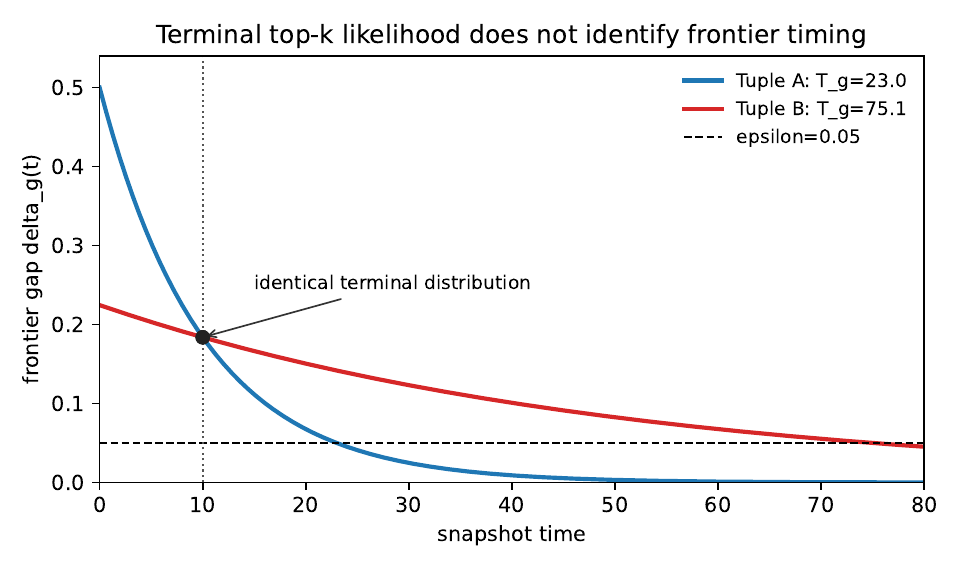}
\caption{Boundary construction used for the observability claim. The terminal selected likelihood is identical at the observed time, but the compatible frontier-gap paths imply \tg\ values of $23.03$ and $75.13$.}
\label{fig:boundary}
\end{figure}

\subsection{Bayesian decision readout}

The same observability boundary also has a decision-theoretic reading. The primitive object is an action chosen to minimize posterior expected loss over a fixed finite action set---not rejection of a null hypothesis. Let $\theta_b$ denote the latent evaluation state for source $b$, including the frontier path, pool and reporting mechanism, benchmark adequacy, and any decision-relevant risk variables. A decision maker observes an archive $D$ and chooses an action
\begin{align}
a \in \mathcal A
=\{&\mathrm{continue},\mathrm{refresh},\mathrm{harden}, \notag\\
&\mathrm{restrict},\mathrm{collect}\}.
\end{align}
Given a loss $L(a,\theta_b)$, the Bayes-optimal action minimizes posterior expected loss \citep{berger1985decision},
\begin{align}
\rho(a\mid D)&=\mathbb E[L(a,\theta_b)\mid D], \notag\\
a^\star(D)&\in\argmin_{a\in\mathcal A}\rho(a\mid D).
\end{align}
In this framing, hypotheses such as open headroom, near plateau, stale evaluation, or unacceptable residual risk are regions of latent state space induced by the action structure: the finite action set $\mathcal{A}$ comes first, and the hypothesis regions are the latent-state partitions that make each action distinguishable, not primitive tests imposed from outside. A posterior-threshold rule is a special case arising from 0-$K_i$ asymmetric loss: if false-positive cost is $c_{10}$ and false-negative cost is $c_{01}$, then the Bayes-optimal action is to act on region $H$ exactly when
\begin{align}
\Pr(H\mid D)>\frac{c_{10}}{c_{10}+c_{01}}.
\end{align}
The threshold $c_{10}/(c_{10}+c_{01})$ is determined entirely by the loss coefficients, not by a universal significance level, and this translation to hypothesis regions is meaningful only after the action and loss structure have been fixed; it does not make the archive protocol a generic Bayesian substitute for null-hypothesis significance testing.

The implemented readout is the finite-draw version of that rule. Each fitted posterior provides draws of $T_b(\epsilon)$, latest headroom, near-threshold and stale-evaluation indicators, residual-risk scores, and headroom shortfall. For each loss family and weight profile, the readout computes
\begin{align}
\widehat\rho(a\mid D)=\frac{1}{M}\sum_{m=1}^{M} L(a,\theta_b^{(m)})
\end{align}
for every action and records both the selected action and all per-action risks. The action set $\mathcal{A}$ and the loss profiles are stylized reporting devices used as scale-controlled diagnostics, not elicited utilities. The \texttt{indicator\_threshold}, \texttt{proxy\_current}, and \texttt{smooth\_residual} families use binary events, the current-gap proxy, and normalized residual/stale-degree scores, respectively, while the balanced, safety-heavy, and staleness-heavy profiles vary the relative cost of missed risk, stale evaluation, over-restriction, refresh, hardening, restriction, and additional collection. The readout therefore characterizes action agreement and regret differences across profiles; the stylized loss structure makes profile-to-profile comparison the appropriate diagnostic unit rather than a single calibrated policy estimate.

The prior enters through the fitted frontier model, not through the decision rule itself. The repeated-snapshot and terminal-only coupled fits use the same dynamic prior family; in the evaluated parameterization, the asymptotic gap is a fraction of $\epsilon$, the initial excess gap is log-normal, the current-headroom fraction is beta-distributed, and the score precision is log-normal. These choices are treated as part of the candidate model under audit. The analysis is scoped to action-level consequences of the observability boundary under this parameterization; prior-sensitivity analysis, model robustness checks, and loss-calibration sweeps lie outside that scope, and the claims below are correspondingly bounded.

The decision readout is an action-facing diagnostic applied to synthetic posterior draws: it surfaces whether the choice of archive type shifts the Bayes-optimal action under the evaluated loss structure. The comparison pairs repeated-snapshot and terminal-only posteriors under three loss families and three weight profiles, yielding $1{,}350$ action comparisons across $150$ paired trajectories. Action-agreement rates are $0.82$ under noisy plateau-like behavior, $0.76$ under candidate-pool growth, and $0.94$ under heavy selective reporting; in the disagreement settings, terminal-under-repeated regret---defined as $\widehat\rho(a_{\mathrm{term}}\mid D)-\widehat\rho(a_{\mathrm{snap}}\mid D)$---is positive, and it is zero when the two regimes select the same action. The observation regime can therefore shift downstream Bayes actions in a fitted model of this type, even when the frontier model itself remains under adjudication.

\section{Candidate Stress Test and Adjudication Protocol}

To exercise the archive-and-adjudication workflow end to end, we evaluate a plausible selection-aware frontier architecture and then subject it to fixed gates. The main candidate, \texttt{S0}, is a dynamic coupled fit over repeated snapshots with a selection-aware likelihood that conditions on the reporting cutoff rather than treating reported leaders as ordinary uncensored draws. Its ablations are defined before use: \texttt{S1} is a static iid fit with neither temporal coupling nor selection correction, \texttt{S2} is static but selection-aware, \texttt{S3} is dynamic but not selection-aware, \texttt{S4} is a deterministic rolling-max heuristic over observed frontier scores, and \texttt{S7} is the same coupled selection-aware architecture as \texttt{S0} conditioned only on the terminal snapshot. The exercise subjects an attractive modeling idea to auditable, falsifiable claims that can be supported, localized, or rejected.

The comparator set is intentionally explicit and minimal. Objective-archive diagnostics also include the latent-effect/factor baseline, while Arena-style preference data uses Bradley-Terry / Elo as the native comparator. Throughout the real-data sections, validation is against future observations rather than latent frontier truth, and Appendix Table~\ref{tab:variant-gates-app} records the same candidate labels together with their gate roles.

The adjudication protocol has six operating components, while Table~\ref{tab:repro-map} expands those claims into seven evidence rows by splitting Section~4 into separate observability-boundary and decision-diagnostic rows. Two conceptually distinct mechanisms appear in that section. The Bayesian decision diagnostics measure action sensitivity under stylized losses, quantifying how posterior beliefs translate into hypothetical decisions across a range of loss parameterizations. The audit gates are falsification-oriented pass criteria applied to model claims: a claim passes only when the stated evidence threshold is met, and the gate yields a localized failure otherwise. The audit gates are not Bayes-optimal decisions in Berger's sense; they are deliberately asymmetric conservative rules motivated by Section~4.1 rather than derived from a decision-theoretic framework, and no operational utility or loss elicitation enters their construction. Gate rules were pre-registered before the reported summaries were inspected. In truth-known recovery, the pre-registered pass criterion requires \texttt{S0} to strictly beat \texttt{S4} and \texttt{S7} on latest-gap error and finite-\tg\ error while not losing to \texttt{S1}--\texttt{S3}; in the slow-frontier negative control, the criterion requires zero false finite-plateau decisions; in the objective backtest, the criterion requires \texttt{S0} to strictly beat both \texttt{S7} and \texttt{S4} on predictive log score using only diagnostic-admissible primary-archive rows and without reversing rank calibration on those rows; and in the calibration audit, only diagnostic-admissible \texttt{S0} rows count, with posterior pass probability at least $0.8$ for the acceptable-calibration region. Here, ``strictly beat'' means a better point summary on the stated metric direction, with ties counted as non-passes, and ``not losing'' means no worse than the reduced comparator on the same admissible metric after the shared filters are applied. The $0/3$ truth-known denominator refers to the three recovery regimes; the slow-frontier setting is a separate negative control. Appendix Table~\ref{tab:variant-gates-app} and Table~\ref{tab:repro-map} map those claims to their audit artifacts, while Table~\ref{tab:main-adjudication} summarizes the resulting evidence.

\begin{table*}[t]
\centering
\small
\setlength{\tabcolsep}{4pt}
\caption{Main archive-and-adjudication readout. Positive rows establish what repeated public snapshots add; diagnostic rows show that the protocol localizes unsupported model claims instead of hiding them.}
\label{tab:main-adjudication}
\begin{tabular}{L{0.22\textwidth}L{0.18\textwidth}L{0.50\textwidth}}
\toprule
Protocol question & Readout & Manuscript consequence \\
\midrule
Can public histories be reconstructed as traces? & Supported & LiveBench and Open LLM Leaderboard v2 become primary objective archives; LMArena is a preference stress test; GAIA is a secondary agentic pilot; tau-bench is an agentic stress-test pilot. \\
Do repeated traces add information? & Supported by construction & Terminal-only top-$k$ evidence can match the selected likelihood while leaving plateau timing materially different in the verified construction, with compatible \tg\ values of $23.03$ and $75.13$. \\
Can traces change action-facing readouts? & Supported diagnostic & In synthetic posterior comparisons, repeated and terminal observation regimes produce loss-sensitive posterior-action disagreements under specified Bayesian decision losses, but not operational policy evidence. \\
Can the archive contract stage aggregate public agentic-result histories? & Supported pilot & GAIA and tau-bench show that the archive contract can stage aggregate public agentic histories, while also exposing missing scaffold and tool metadata. \\
Can the protocol reject unsupported models? & Supported diagnostic & The candidate method does not earn truth-known recovery, primary-archive prediction, preference transfer, or calibrated timing uncertainty. \\
\bottomrule
\end{tabular}
\end{table*}

\begin{table*}[t]
\centering
\scriptsize
\setlength{\tabcolsep}{2pt}
\caption{Claim-to-artifact reproducibility map for the headline manuscript evidence. The map includes Section~4 boundary and decision-diagnostic artifacts in addition to the six model-adjudication protocol components.}
\label{tab:repro-map}
\begin{tabularx}{\linewidth}{@{}>{\raggedright\arraybackslash}p{2.4cm} >{\raggedright\arraybackslash}p{4.8cm} >{\raggedright\arraybackslash}p{4.3cm} >{\raggedright\arraybackslash}X@{}}
\toprule
Claim & Main artifact & Driver / readout & Audit use \\
\midrule
Truth-known recovery & Synthetic gate summary and machine-readable companion. & Frontier metric summary and gate readout. & Synthetic recovery regimes, slow-frontier negative control, and $0/3$ gate readout. \\
Archive validation & Archive validation table. & Archive build summary. & Source grades, fold counts, duplicate policy, and exclusion reasons. \\
Observability boundary & Terminal-boundary verification and Figure~\ref{fig:boundary}. & Boundary evidence index. & Shared terminal likelihood and differing \tg\ values. \\
Objective backtest & Objective headline table and paired bootstrap intervals. & Rolling-origin manifest. & Primary-archive predictive diagnostics and comparator rule. \\
Preference stress test & Arena headline table and paired bootstrap intervals. & Arena stress summary. & LMArena stress-test diagnostics against \texttt{S7} and \texttt{BT/ELO}. \\
Decision diagnostic & Exact observation-regime summary. & Exact decision-readout driver. & Repeated-vs-terminal action agreement and regret summaries. \\
Calibration audit & SBC report and calibration failure table. & Coverage-by-regime and diagnostic-pathology summaries. & Acceptable-calibration posterior probabilities and interval pathologies. \\
\bottomrule
\end{tabularx}
\end{table*}

\section{Results}

We report the results as separate checks rather than collapsing them into a single aggregate score, because the evaluation measures distinct claims. Archive validity, observability, decision relevance, agentic applicability, predictive adequacy, preference transfer, and posterior calibration do different jobs and can succeed or fail independently. The archive-and-adjudication contribution is supported, while the candidate model does not meet the predictive and calibration criteria.

\begin{figure*}[t]
\centering
\includegraphics[width=\textwidth]{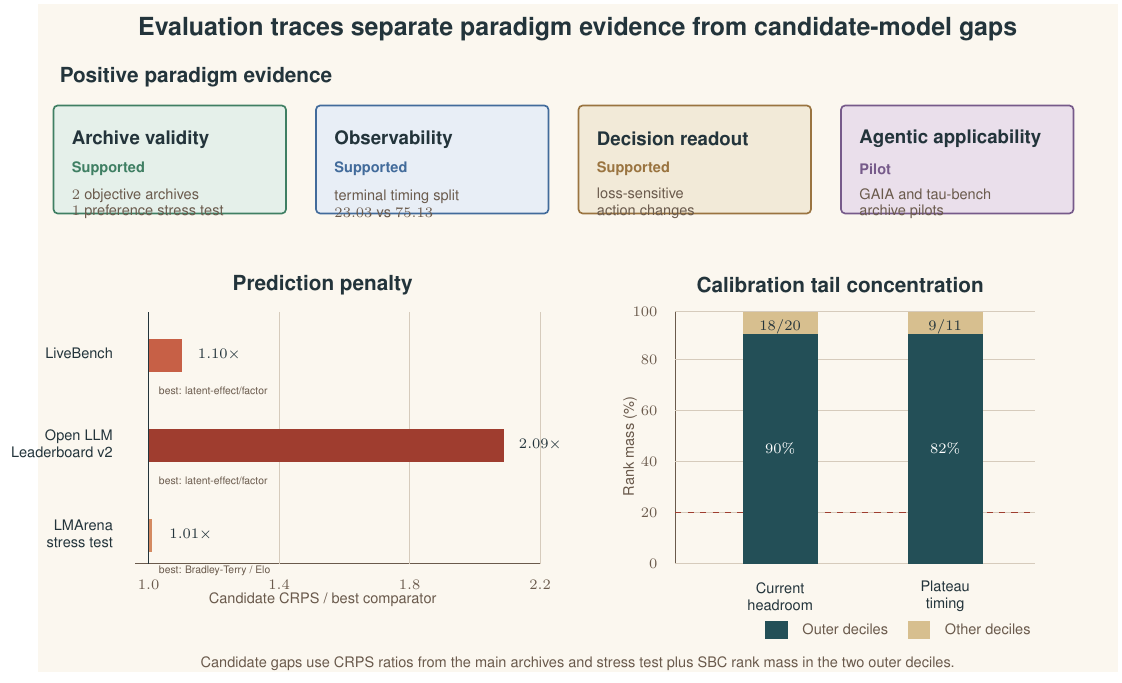}
\caption{What the trace-adjudication protocol separates beyond terminal tables. The positive evidence is archive validity, observability, decision relevance, and agentic applicability; the candidate-model gaps are predictive shortfalls and posterior calibration failure.}
\label{fig:result-insight-map}
\end{figure*}

\paragraph{Technical definitions.}
Several terms recur in this section in a narrow technical sense. Predictive log score is the average held-out predictive log likelihood,
\begin{align}
\operatorname{LogScore} &= \frac{1}{n}\sum_{r=1}^n \log p(y_r^{\mathrm{hold}}\mid D_r^{\mathrm{train}}),
\end{align}
so higher is better. \mbox{Continuous ranked probability score} (CRPS) is the proper score
\begin{align}
\operatorname{CRPS}(F,y) &= \int_{-\infty}^{\infty}\left(F(z)-\mathbf{1}\{y\le z\}\right)^2\,dz,
\end{align}
so lower is better. For the Normal predictive distributions used in the backtest metric, if $F=\mathcal{N}(\mu,\sigma^2)$ and $u=(y-\mu)/\sigma$, then
\begin{align}
\operatorname{CRPS}(F,y)
  &= \sigma\Bigl[u\bigl(2\Phi(u)-1\bigr) \notag\\
  &\qquad + 2\phi(u)-\frac{1}{\sqrt{\pi}}\Bigr].
\end{align}
Item response theory (IRT) motivates a common latent-ability form, which provides context for the shrinkage baseline used as the objective comparator,
\begin{align}
\Pr(Y_{iq}=1) &= \sigma\!\left(a_q(\theta_i-\beta_q)\right), \notag\\
\sigma(x) &= \frac{1}{1+e^{-x}},
\end{align}
where system ability $\theta_i$, item difficulty $\beta_q$, and discrimination $a_q$ parameterize the response. In this paper, the objective-archive comparator is a shrinkage latent-effect/factor baseline that separates system/model, family, task, and time effects, for example through a linear predictor of the form
\begin{align}
\eta_{mfqt} &= \mu + \alpha_m + \gamma_f + \delta_q + \tau_t,
\end{align}
with partial pooling on those components. This baseline serves as the diagnostic comparator for objective archives. The terminal-history baseline conditions only on the final coupled evidence, while the rolling-frontier heuristic extrapolates a simple time-indexed frontier without the candidate model's selection-aware likelihood.

For the preference stress test, LMArena denotes leaderboard rating snapshots derived from Arena-style preference evaluations rather than an objective archive. Bradley-Terry uses
\begin{align}
\Pr(i \succ j) &= \sigma(\eta_i-\eta_j),
\end{align}
and Elo is a rating and update variant built on the same latent-strength idea. Bradley-Terry / Elo therefore serves as the native comparator for preference-style ratings. Top-$k$ recall is
\begin{align}
R_k(y,\hat y) &= \frac{|S_k(y)\cap S_k(\hat y)|}{k},
\end{align}
where $S_k(\cdot)$ is the reported top-$k$ set. Rank calibration is the Spearman correlation between observed and predicted descending ranks,
\begin{align}
\operatorname{RankCal}(y,\hat y) &= \rho_{\mathrm{S}}\!\left(r^\downarrow(y),\,r^\downarrow(\hat y)\right).
\end{align}
Paired cluster-bootstrap intervals report primary-minus-comparator metric differences under resampling over fold clusters.

Simulation-based calibration (SBC) uses the posterior rank or midrank of the data-generating truth among posterior draws,
\begin{align}
R &= \sum_{m=1}^M \mathbf{1}\{\theta^{(m)}<\theta^\star\} + \frac{1}{2}\sum_{m=1}^M \mathbf{1}\{\theta^{(m)}=\theta^\star\},
\end{align}
which is uniform when the posterior is calibrated. If these SBC ranks are binned into $B$ bins with counts $n_1,\ldots,n_B$, the Dirichlet rank-bin posterior is
\begin{align}
p \mid n &\sim \operatorname{Dirichlet}(\alpha+n_1,\ldots,\alpha+n_B).
\end{align}
The reported audit uses the symmetric setting $\alpha=1.0$. The Kolmogorov-Smirnov and chi-square summaries used later measure posterior discrepancy over those rank-bin probabilities,
\begin{align}
D_{\mathrm{KS}}(p) &= \max_{1\le b\le B}\left|\sum_{j=1}^b p_j-\frac{b}{B}\right|, \\
D_{\chi^2}(p) &= \sum_{b=1}^B \frac{(p_b-1/B)^2}{1/B}.
\end{align}

Truth-known recovery is $0/3$, objective-archive prediction is $0/2$, the Arena stress test shows a preference-transfer gap, and posterior calibration fails for both current headroom and finite timing. Tables~\ref{tab:objective-main} and~\ref{tab:arena-main} give the predictive details, while Figure~\ref{fig:result-insight-map} summarizes what the trace-adjudication protocol separates beyond terminal tables.

\subsection{Truth-known admissibility}

The truth-known check shows why model gates must be fixed before real-archive performance is interpreted as evidence about latent frontiers. Across controlled settings for closing gaps, noisy plateau-like behavior, and changes in the candidate pool, the candidate method fails to improve on the terminal and rolling comparators while avoiding losses to reduced variants. The slow-frontier negative control behaves correctly, confirming that the method does not produce a false plateau there. Synthetic evidence is therefore not an optional appendix at the archive level; it is the first place a frontier-inference method must pass before real-data claims can be interpreted.

\subsection{Primary archive prediction}

The objective-archive check tests future-observation prediction on the two validated primary archives: LiveBench and Open LLM Leaderboard v2. LMArena is kept out of this check because it is a preference stress test, while excluded sources appear in Appendix~\ref{app:tables}.

This prediction check is a positive result for the archive protocol, but not for the candidate method. On LiveBench, the candidate trails both the terminal-history baseline and the latent-effect/factor diagnostic baseline on predictive log score and CRPS. On Open LLM Leaderboard v2, it is close to the terminal-history baseline on log score but still worse on both log score and CRPS, while the latent-effect/factor baseline is much stronger. The absolute log-score magnitudes are far more extreme than the CRPS scale and remain a metric-scale or tail-dominance audit item; the table therefore supports only the directional gate readout, not a substantive interpretation of the log-score scale. The rolling-frontier heuristic is weaker than the candidate on these aggregate metrics, but the adjudication rule requires the candidate to beat the required comparators on enough primary archives, and it does not. The key finding is that repeated public traces support rolling-origin prediction tests with explicit comparators rather than terminal leaderboard snapshots.

\begin{table*}[t]
\centering
\scriptsize
\setlength{\tabcolsep}{3pt}
\caption{Objective-archive predictive details. Higher log score and lower CRPS are better; candidate gaps are localized against explicit comparators. The very large log-score magnitudes are retained from the audit artifact but should be read as directional until the score scale and tail influence are separately audited.}
\label{tab:objective-main}
\begin{tabular}{L{0.24\textwidth}L{0.28\textwidth}rrL{0.22\textwidth}}
\toprule
Archive & Method & Log score & CRPS & Protocol readout \\
\midrule
LiveBench & Latent-effect/factor baseline & $-9.67 \times 10^9$ & $0.128$ & Best comparator \\
LiveBench & Terminal-history baseline & $-1.06 \times 10^{10}$ & $0.133$ & Required comparator \\
LiveBench & Selection-aware candidate & $-1.24 \times 10^{10}$ & $0.141$ & Candidate gap localized \\
LiveBench & Rolling-frontier heuristic & $-4.24 \times 10^{10}$ & $0.282$ & Weaker heuristic \\
\midrule
Open LLM Leaderboard v2 & Latent-effect/factor baseline & $-1.28 \times 10^{12}$ & $6.136$ & Best comparator \\
Open LLM Leaderboard v2 & Terminal-history baseline & $-3.95 \times 10^{12}$ & $12.558$ & Required comparator \\
Open LLM Leaderboard v2 & Selection-aware candidate & $-3.99 \times 10^{12}$ & $12.821$ & Candidate gap localized \\
Open LLM Leaderboard v2 & Rolling-frontier heuristic & $-1.19 \times 10^{13}$ & $22.659$ & Weaker heuristic \\
\bottomrule
\end{tabular}
\end{table*}

Paired cluster-bootstrap intervals sharpen the same diagnosis while preserving that scale caveat. Against the latent-effect/factor baseline, the candidate-minus-baseline intervals are negative for log score ($[-3.00 \times 10^{12}, -2.17 \times 10^{11}]$) and positive for CRPS ($[3.24, 3.94]$), so both summaries point against the candidate directionally. Against the terminal-history baseline, the log-score interval crosses zero ($[-5.56 \times 10^{10}, 9.29 \times 10^9]$), but the CRPS interval remains positive ($[0.085, 0.217]$). Table~\ref{tab:objective-main} reports the full archive-by-method details including the rolling-frontier heuristic. The objective archives validate the archive testbed while localizing the candidate's predictive gap, confirming the protocol's intended separation between archive value and model endorsement.

\subsection{Preference-regime stress test}

Arena-style preference data defines a different measurement regime, because prompt mix, population, judge, release timing, and access confounds are not interchangeable with objective benchmark scores. LMArena is therefore used as a stress test rather than as a primary objective archive. On the main Arena comparison, the candidate trails the shared terminal-history / Bradley-Terry-Elo readout on predictive log score, CRPS, top-$k$ recall, and rank calibration, with resampling intervals pointing the same way. In this snapshot-derived stress test, the terminal-history and Bradley-Terry / Elo rows coincide numerically, so they reflect one underlying comparison reported under two naming conventions rather than two independent pieces of evidence. Preference-transfer claims therefore need source-specific adjudication rather than a generic frontier narrative.

\begin{table}[t]
\centering
\scriptsize
\setlength{\tabcolsep}{2pt}
\caption{Preference-stress-test details. Higher log score, top-$k$ recall, and rank calibration are better; lower CRPS is better. The first two rows are numerically identical in this snapshot-derived stress test and should not be counted as independent evidence.}
\label{tab:arena-main}
\begin{tabular}{L{0.31\linewidth}rrrr}
\toprule
Method & Log score & CRPS & Top-$k$ & Rank cal. \\
\midrule
Bradley-Terry / Elo & $-4.3190$ & $7.4769$ & $0.8094$ & $0.9958$ \\
Terminal-history baseline (same readout here) & $-4.3190$ & $7.4769$ & $0.8094$ & $0.9958$ \\
Selection-aware candidate & $-4.3250$ & $7.5581$ & $0.8047$ & $0.9957$ \\
\bottomrule
\end{tabular}
\end{table}

Candidate-minus-Bradley-Terry / Elo cluster-bootstrap intervals are in the wrong direction for every reported metric: log score $[-0.00793, -0.00410]$, CRPS $[0.0569, 0.1065]$, top-$k$ recall $[-0.00842, -0.00101]$, and rank calibration $[-1.45 \times 10^{-4}, -2.04 \times 10^{-5}]$. Table~\ref{tab:arena-main} gives the compact metric readout. The preference stress test provides no support for transfer claims, even though the aggregate metric differences are numerically modest.

\subsection{Posterior uncertainty audit}

\begin{figure}[t]
\centering
\includegraphics[width=\linewidth]{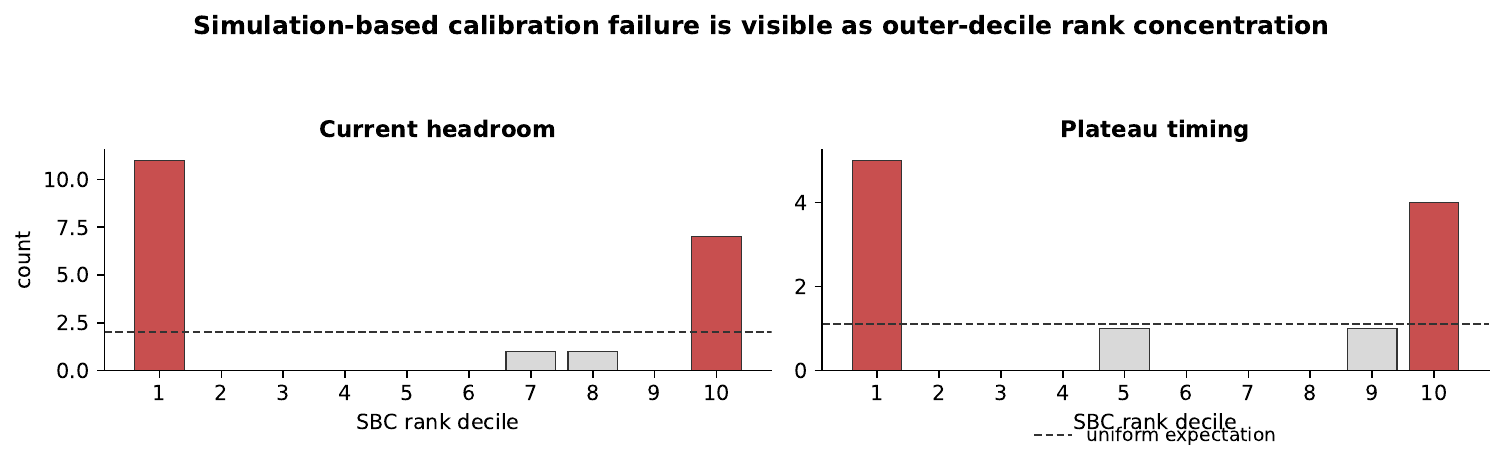}
\caption{Simulation-based calibration ranks concentrate in the outer deciles: current headroom has $18/20$ admissible ranks with posterior acceptable-calibration probability $1 \times 10^{-5}$; finite timing has $9/11$ outer-decile ranks with probability $0.0133$.}
\label{fig:sbc-ranks}
\end{figure}

The calibration audit shows how the protocol handles posterior uncertainty claims. Using the available posterior draws, simulation-based calibration ranks are computed and a Dirichlet posterior is placed on the $10$-bin rank histogram with symmetric concentration $\alpha=1.0$; posterior mass is then evaluated over an acceptable-calibration region defined by total-variation distance to uniform at most $0.35$, outer-decile mass between $0.10$ and $0.30$, and overall posterior pass probability at least $0.8$. Current-headroom SBC places $18/20$ admissible ranks in the outer deciles and gives posterior acceptable-calibration probability $1 \times 10^{-5}$. Finite-timing SBC places $9/11$ admissible ranks in the outer deciles and gives posterior acceptable-calibration probability $0.0133$. These summaries rest on only $20$ and $11$ admissible ranks, respectively, so they are small-sample diagnostics rather than definitive calibration verdicts; nevertheless, the observed outer-decile masses, $0.90$ and $0.82$, are far outside the acceptable interval $[0.10,0.30]$. Figure~\ref{fig:sbc-ranks} shows this rank concentration directly.

The KS- and chi-square-style summaries are secondary posterior discrepancy checks over rank-bin probabilities, not null-hypothesis significance tail probabilities. They are reported as descriptive posterior functionals of the same rank-bin distribution used by the primary acceptable-calibration gate. For this candidate, they do not override the primary failure, and the timing pathologies remain substantively relevant because \tg\ is a thresholded path functional: when the local slope is weak near the threshold, small posterior shifts can sharply move the inferred crossing time, collapse an interval, or push posterior mass onto effectively infinite branches. That pattern is not consistent with calibrated timing inference for this candidate method.

\section{Conclusion}

The archive-and-adjudication protocol produces three concrete results. Repeated public snapshots normalize into graded archives; terminal-only evidence yields a verified counterexample to timing identification under a fixed reporting convention; and Bayesian posterior expected-loss diagnostics vary by observation regime in synthetic comparisons. The archive contract extends to GAIA and tau-bench as limited-applicability pilots, exposing missing scaffold and tool metadata as a boundary condition of the observability standard.

The candidate model fails all four falsification gates---synthetic recovery, objective backtesting, Arena preference transfer, and calibration---demonstrating that the protocol separates models that meet its evidentiary standards from those that do not. The transparent record and adjudication structure tie public evaluation claims to verifiable evidence, so that any evaluation claim submitted against this protocol is either supported or falsified by the archive rather than left to assertion.

\clearpage

\section*{Limitations}

Several limitations are structural rather than accidental. Real archives provide future-observation validation rather than latent frontier truth, and candidate-pool reconstruction remains partly assumption-driven. The current candidate model uses a simple monotone gap family, with no demonstrated robustness to power-law, stretched-exponential, or non-monotone frontier dynamics. The source-normalized law $F_{bt}^{\star}$, ceiling $u_b$, and timing threshold $\epsilon=0.05$ are model-and-reporting conventions, not objects recovered without assumptions from selected top-$k$ records. The Bayesian decision layer uses stylized loss families and synthetic posterior draws; its scope is diagnostic sensitivity of action thresholds to prior and loss specification, not validated governance policy. Because no prior, model, or loss robustness sweep is performed, posterior timing and action readouts reflect the specific conventions chosen and should not be generalized as operational prescriptions. Moreover, Berger-style posterior summaries are decision-relevant only relative to the loss under which they are reported: without elicited utilities, the loss profiles used here are diagnostic conventions rather than operationally calibrated policy inputs. LiveBench scores reflect a dated model-judgment aggregate, which constrains the recency and generalizability of any benchmark-derived saturation signal. Preference archives remain confounded by sampling and access, and excluded archives sit outside the evidence baseline; as a result, conclusions about saturation timing for those archives cannot be drawn from the present data. The GAIA and tau-bench pilots show archive applicability for agentic-style histories, not full agent-trace observability, since richer agentic traces still need tool budget, scaffold identity, environment version, retry policy, judge version, and human-intervention fields.

\appendix

\section{Detailed Evidence}
\label{app:tables}

This appendix collects the compact evidence needed to audit the main-text claims in one place. Table~\ref{tab:archive-app} expands the archive-source inventory, Table~\ref{tab:variant-gates-app} defines the candidate variants and gate uses, Table~\ref{tab:gates-app} summarizes the gate outcomes, and Table~\ref{tab:agentic-app} reports the agentic applicability pilots.

\begin{table*}[!t]
\centering
\small
\caption{Expanded source inventory for the public evaluation archive.}
\label{tab:archive-app}
\begin{tabularx}{\textwidth}{@{}>{\raggedright\arraybackslash}p{2.8cm} >{\raggedright\arraybackslash}p{2.5cm} r r r >{\raggedright\arraybackslash}X@{}}
\toprule
Source & Public role & Snapshots & Systems & Validation slices & Notes \\
\midrule
LiveBench & Primary objective archive & $94$ & $195$ & $91$ & Dated model-judgment aggregate; source-native timestamps; top-$k$ reconstructable. \\
Open LLM Leaderboard v2 & Primary objective archive & $262$ & $4484$ & $259$ & Submission-date snapshots; flagged rows removed; rank reconstructed from score ordering. \\
LMArena leaderboard snapshots & Preference stress test & $152$ & $365$ & $149$ & Preference archive kept separate from objective headline claims. \\
GAIA public results & Secondary agentic pilot & $463$ & $3353$ & $460$ & Aggregate agentic-style public rows; dated level scores are usable, scaffold and tool metadata are weak. \\
tau-bench public submissions & Agentic stress-test pilot & $10$ & $27$ & $7$ & Agentic tool-use submissions with domain, modality, retrieval/voice, and Pass@k metadata; kept outside main objective claims. \\
LiveCodeBench & Excluded & $0$ & $0$ & $0$ & Versioned source table unavailable in the public histories used here. \\
HELM Capabilities & Excluded & $0$ & $0$ & $0$ & Versioned source table unavailable in the public histories used here. \\
SWE-bench Verified & Excluded & $0$ & $0$ & $0$ & Versioned source table unavailable in the public histories used here. \\
\bottomrule
\end{tabularx}
\end{table*}

\begin{table*}[!t]
\centering
\small
\caption{Candidate variants and their gate roles.}
\label{tab:variant-gates-app}
\begin{tabularx}{\linewidth}{@{}l >{\raggedright\arraybackslash}p{5.0cm} >{\raggedright\arraybackslash}X@{}}
\toprule
Label & Construction & Gate role \\
\midrule
\texttt{S0} & Selection-aware dynamic coupled fit over repeated snapshots. & Retained only as a failing stress-test object for adjudication; not an endorsed archive variant. \\
\texttt{S1} & Static iid fit without temporal coupling or selection correction. & Reduced-variant non-loss comparator in truth-known recovery. \\
\texttt{S2} & Static selection-aware fit without temporal coupling. & Reduced-variant non-loss comparator in truth-known recovery. \\
\texttt{S3} & Dynamic fit without selection correction. & Reduced-variant non-loss comparator in truth-known recovery. \\
\texttt{S4} & Deterministic rolling-max heuristic over observed frontier scores. & Required strict-win comparator in truth-known recovery and objective backtests. \\
\texttt{S7} & Same coupled selection-aware architecture as \texttt{S0}, but conditioned only on the terminal snapshot. & Required strict-win comparator in truth-known recovery; terminal-history comparator in objective backtests and decision diagnostics. \\
\texttt{BT/ELO} & Bradley-Terry / Elo preference comparator applied to Arena-style leaderboard rating snapshots. & Native comparator for the LMArena stress test. \\
\bottomrule
\end{tabularx}
\end{table*}

\begin{table*}[!t]
\centering
\small
\caption{Adjudication summary for the evaluated evidence.}
\label{tab:gates-app}
\begin{tabularx}{\linewidth}{@{}l l >{\raggedright\arraybackslash}X@{}}
\toprule
Check & Status & Key readout \\
\midrule
Truth-known synthetic recovery & not supported & The candidate method passed $0/3$ recovery regimes; only the slow-frontier negative control behaved correctly. \\
Archive validation & supported & Two objective archives validated as primary evidence; one preference archive validated as a stress test; GAIA validated as a secondary agentic pilot; tau-bench validated as an agentic stress-test pilot. \\
Objective backtest & not supported & The candidate method passed $0/2$ primary archives under the fixed comparator rule. \\
Preference stress test & not supported & The candidate method trails the shared terminal-history / Bradley-Terry-Elo readout on the main Arena comparison. \\
Observability boundary & supported by construction & Terminal-only evidence can match the selected likelihood while yielding \tg\ values $23.03$ and $75.13$ in the verified counterexample. \\
Bayesian decision readout & supported diagnostic & Synthetic posterior comparisons show loss-sensitive repeated-vs-terminal action differences, but not operational superiority. \\
Calibration audit & not supported & Simulation-based calibration gives low posterior probability to acceptable calibration for the candidate model; the interval audit finds missed finite-time intervals, degenerate intervals, effectively infinite timing intervals, and numerical instabilities. \\
\bottomrule
\end{tabularx}
\end{table*}

\begin{table*}[!t]
\centering
\small
\caption{Agentic archive applicability pilots. These rows demonstrate archive applicability, not candidate-model superiority.}
\label{tab:agentic-app}
\begin{tabularx}{\textwidth}{@{}>{\raggedright\arraybackslash}p{3.0cm} >{\raggedright\arraybackslash}p{2.4cm} r r r r >{\raggedright\arraybackslash}X@{}}
\toprule
Source & Public role & Snapshots & Systems & Task groups & Diagnostic rows & Interpretation \\
\midrule
GAIA public results & Secondary agentic pilot & $463$ & $3353$ & $8$ & $11784$ & Completed aggregate public-result applicability pilot; scaffold and tool metadata remain weak. \\
tau-bench public submissions & Agentic stress-test pilot & $10$ & $27$ & $3$ & $27$ & Completed aggregate public-result applicability pilot over voice Pass@$1$ slices; richer submission metadata, but not a main objective archive. \\
\bottomrule
\end{tabularx}
\end{table*}

\end{document}